\documentclass[twocolumn]{article}

\usepackage{geometry}
\usepackage{authblk}
\usepackage{cite}
\usepackage[utf8]{inputenc}
\usepackage{multirow}
\usepackage{graphicx}
\usepackage{outlines}
\usepackage[dvipsnames]{xcolor}
\usepackage{textcomp}
\usepackage{float}
\usepackage{ulem}
\usepackage{comment}
\usepackage{amssymb}
\usepackage{mathtools}
\usepackage[dvipsnames]{xcolor}

\newcommand{\onlinecite}[1]{\hspace{-1 ex} \nocite{#1}\citenum{#1}}

\usepackage{hyperref}
\hypersetup{
    citecolor = black,
    filecolor = black,
    urlcolor = black,
    colorlinks = true, 
    linktoc = all,     
    linkcolor = black,  
}

\begin{document}

\title{An active dendritic tree can mitigate fan-in limitations\\in superconducting neurons}
\vspace{-0.5em}
\author[1,2]{Bryce A. Primavera}
\author[2]{Jeffrey M. Shainline}
\affil[1]{Department of Physics, University of Colorado Boulder}
\affil[2]{National Institute of Standards and Technology}
\date{\today}

\twocolumn[
\begin{@twocolumnfalse}
\maketitle
\begin{abstract}
Superconducting electronic circuits have much to offer with regard to neuromorphic hardware. Superconducting quantum interference devices (SQUIDs) can serve as an active element to perform the thresholding operation of a neuron's soma. However, a SQUID has a response function that is periodic in the applied signal. We show theoretically that if one restricts the total input to a SQUID to maintain a monotonically increasing response, a large fraction of synapses must be active to drive a neuron to threshold. We then demonstrate that an active dendritic tree (also based on SQUIDs) can significantly reduce the fraction of synapses that must be active to drive the neuron to threshold. In this context, the inclusion of a dendritic tree provides the dual benefits of enhancing the computational abilities of each neuron and allowing the neuron to spike with sparse input activity.
\vspace{2em}
\end{abstract}
\end{@twocolumnfalse}
]


\section{\label{sec:introduction}Introduction}
Motivations for developing artificial spiking neural systems include efficient hardware implementations of brain-inspired algorithms and construction of large-scale systems for studying the mechanisms of cognition. While most efforts toward these ends employ semiconductor hardware based on silicon transistors \cite{lide2015,indiveri2011neuromorphic,merolla2014million,davies2018loihi,scpo2017}, superconducting electronics have also received considerable attention. Superconducting circuits based on Josephson junctions (JJs, \cite{vatu1998,ka1999}) have strengths that make them appealing for neural systems, including high speed, low energy consumption per operation, and native thresholding/spiking behaviors. In particular, two-junction superconducting quantum intereference devices (SQUIDs) are ubiquitous in superconducting electronics, and several efforts aim to utilize SQUIDs for various neuromorphic operations \cite{hago1991,hiak1991,mina1994,mina1995,rilo1997,crsc2010,onma2011,chca2013,yaum2013,sele2017,kafu2018,sh2018,scdo2018,scdo2018b,tose2019}. 

\begin{figure*}[t!]
\centering
\includegraphics[width=17.2cm]{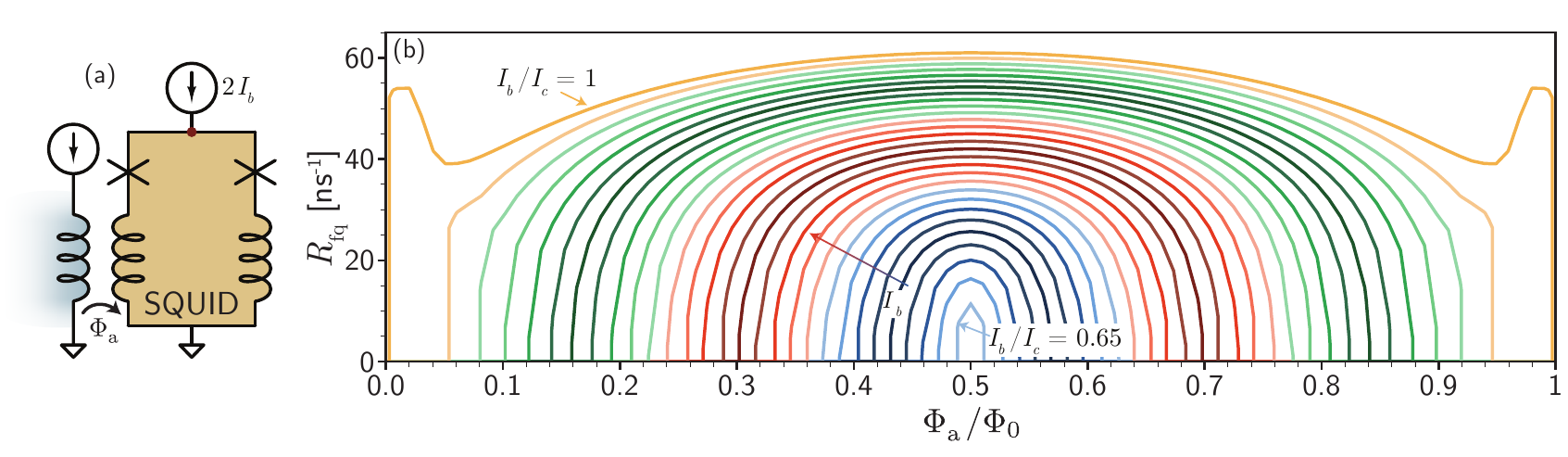}
\caption{(a) SQUID circuit with DC current bias ($I_b$ through each JJ) and flux input ($\Phi_{\mathrm{a}}$) through a transformer. (b) SQUID response function. $R_{fq}$ is the rate of flux-quantum production due to the voltage developed at the node indicated by a red dot in part (a) as a function of the applied flux to the SQUID loop in units of the magnetic flux quantum $\Phi_0 = h/2e$. Different curves correspond to different bias conditions.}
\label{fig:squid_response}
\end{figure*}

As shown in Fig.\,\ref{fig:squid_response}(a), a SQUID has two inputs: a bias current (typically DC) that sets the operating point of the device and a flux input coil. The response of the SQUID to the flux input is peculiar in that it is periodic, as shown in Fig.\,\ref{fig:squid_response}(b). One way to use a SQUID as a neuromorphic component is to collect the inputs from many synapses or dendrites and apply them as flux to the SQUID loop. The response then has a threshold that depends on the applied bias current, $I_b$, and is maximal at an applied flux of $\Phi_0/2$, where $\Phi_0 = h/2e \approx 2$\,mV$\cdot$ps is the magnetic flux quantum. In this work, we consider the ramifications of limiting the maximum applied flux to this value of $\Phi_0/2$ so the response is monotonic with applied signal.

Fan-in has recently been analyzed in superconducting neuromorphic circuits wherein single-flux quanta are used as signals between neurons \cite{schneider2020fan}. However, that work was not concerned with the case in which analog synaptic signals integrated and stored over time could drive a SQUID beyond the first half period of its response function. In the present study, we analyze fan-in in the context of leaky-integrator neuronal circuits that were originally designed for use in large-scale superconducting optoelectronic systems \cite{shbu2017,sh2019,sh2020,sh2021}. However, the conclusions of this paper should be applicable to a wide variety of SQUID neurons.

In this work, we use the following component definitions. A \textit{synapse} is a circuit that receives a single input from another neuron and produces an electrical current circulating in a storage loop. A \textit{dendrite} is a circuit that receives an input proportional to the electrical output of one or more synapses and/or dendrites, performs a transfer function on the sum of the inputs, and produces an electrical current circulating in a storage loop as the output. A \textit{neuron cell body} (also known as a \textit{soma}) receives input proportional to the electrical output of one or more synapses and/or dendrites, performs a threshold operation on the sum of the inputs, and produces an output pulse if the threshold is exceeded. Outputs from the neuron cell body are routed to many downstream synapses. \textit{Fan-in} is the collection and localization of multiple synaptic or dendritic signals into a dendrite or neuron cell body. 

\section{\label{sec:model}Model}
We begin by considering the simple SQUID circuit shown in Fig.\,\ref{fig:squid_response}(a). The weighted sum of synaptic signals is represented by the applied flux ($\Phi_\mathrm{a}$) to the SQUID loop. Upon reaching a threshold value of applied flux ($\Phi_\mathrm{a}^\mathrm{th}$), the SQUID produces a train of fluxons.  Fig.\,\ref{fig:squid_response}(b) shows the relationship between the rate of fluxon production ($R_\mathrm{fq}$) and $\Phi_\mathrm{a}^\mathrm{th}$. $\Phi_\mathrm{a}^\mathrm{th}$ depends on $I_b$ and corresponds to the $x$-intercepts in the figure. The generated fluxon train can be used in multiple ways depending on the neuromorphic context. For instance, each fluxon produced could be interpreted as an entire action potential, for use in either rate or temporal coding schemes \cite{scdo2018b, sele2017}. Alternatively, the rate of fluxon production could be treated as an analog output that triggers another thresholding device (such as an optical transmitter) that drives action potentials downstream \cite{sh2019}.  In any of these contexts, however, maintaining a monotonic response is desirable. We will first treat the point-neuron case (synapses are connected directly to the neuron cell body), discuss its limitations, and then investigate how an active dendritic arbor mitigates these concerns.

\subsection{\label{sec:point_neuron}Point neurons with a SQUID soma}
As seen in Fig.\,\ref{fig:squid_response}(b), restricting the applied flux to the range $0 \le \Phi_{\mathrm{a}} \le \Phi_0/2$ would enforce monotonicity. However, restricting the signal requires a large fraction of synapses to be active to reach threshold. The largest signal that can be applied to the SQUID occurs when all $n$ synapses are active and maximally weighted. Suppose a maximally weighted synapse applies $\Phi_\mathrm{sy}$ of flux to the SQUID. To ensure monotonicity, we set this maximum possible signal equal to $\Phi_0/2$:
\begin{equation}
    n\Phi_\mathrm{sy} = \frac{\Phi_0}{2}.
\end{equation}
To reach threshold, a critical number $p$ of synaptic inputs must be active. For simplicity, we assume that each active input supplies $\Phi_\mathrm{sy}$ of flux. If the applied flux necessary to reach threshold is $\Phi_\mathrm{a}^\mathrm{th}$, then
\begin{equation}
\label{eq:delta_phi}
    \Phi_\mathrm{a}^\mathrm{th} = p\,\Phi_\mathrm{sy}  = \frac{p\,\Phi_0}{2n}.
\end{equation}
At threshold, the induced current is equal to the difference between $I_\mathrm{c}$ and $I_\mathrm{b}$, where $I_\mathrm{b}$ is the current through one of the JJs when no flux is applied to the loop. For a SQUID with total inductance $L_{\mathrm{tot}}^{\mathrm{sq}}$, this implies that $\Phi_\mathrm{a}^\mathrm{th} = L^{\mathrm{sq}}_{\mathrm{tot}}(I_\mathrm{c} - I_\mathrm{b})$. We can rewrite Eq.\,\ref{eq:delta_phi} in terms of the critical and bias currents and rearrange to find the minimum fraction of synapses that must be active to reach threshold:
\begin{equation}
    \frac{p}{n} = \frac{2L^{\mathrm{sq}}_{\mathrm{tot}}(I_{\mathrm{c}}-I_{\mathrm{b}})}{\Phi_0}.
\end{equation}
 $L^{\mathrm{sq}}_{\mathrm{tot}}$ is a function of flux in the SQUID through the variable inductance of JJs. For a typical SQUID near threshold, $L^{\mathrm{sq}}_{\mathrm{tot}} \approx \frac{\Phi_0}{I_{\mathrm{c}}}(\frac{3\pi + 2}{4\pi})$ (see Appendix \ref{apx:squid}), resulting in the expression
\begin{equation}
\label{eq:fraction_active}
\frac{p}{n} = \frac{3\pi + 2}{2\pi} \left( 1-\frac{I_b}{I_c} \right).
\end{equation}

Equation \ref{eq:fraction_active} gives the fraction of maximally weighted synapses that must be active to drive the SQUID to threshold. When one limits the total signal that can be applied to a SQUID to $\Phi_0/2$ to maintain a monotonic response, the threshold activity fraction is dependent only on the ratio $I_b/I_c$. Noise prohibits biasing with $I_b$ arbitrarily close to $I_c$. The effects of noise in superconducting neurons warrant further investigation, but $I_{\mathrm{b}}/I_{\mathrm{c}} = 0.7$ is typical of digital superconducting systems. This value corresponds to a minimum threshold activity of about 55\%. $I_\mathrm{b}/I_\mathrm{c} = 0.9$ would represent an aggressive operating point, and would require $p/n \approx 18\%$. Such an activity level is higher than that observed in biology, where 1\% - 16\% of neurons may be active at any time due to power considerations as well as implications of sparse coding \cite{laughlin2003communication}. As stated in Ref.\,\onlinecite{laughlin2003communication}, ``Sparse coding schemes, in which a small proportion of cells signal at any one time, use little energy for signaling but have high representational capacity, because there are many different ways in which a small number of signals can be distributed among a large number of neurons.'' We next consider how performing fan-in with an active dendritic tree comprised of SQUIDs alters the calculation.

\subsection{\label{sec:dendritic_tree}Fan-in with an active dendritic tree}
Biological neurons are not accurately modeled as point neurons. Instead, synaptic inputs are passed through an arbor of active dendrites that performs numerous computations \cite{mel1994information,loha2005,stsp2015}, including intermediate threshold functions between subsets of synapses and the soma \cite{sava2017} and detection of synaptic sequences \cite{haah2016}. Active dendrites can be significant for adaptation and plasticity \cite{majo2005,sjostrom2008dendritic}, can dramatically increase information storage capacity relative to point neurons \cite{poirazi2001impact},  and when modulated by inhibitory neurons, the dendritic tree can induce a given neuron to perform distinct computations at different times, enabling a given structural network to dynamically realize myriad functional networks \cite{bu2006}. Discussion of dendrites in superconducting neural hardware is found in Ref. \onlinecite{sh2020}. 

\begin{figure}[htb]
\centering
\includegraphics[width=8.6cm]{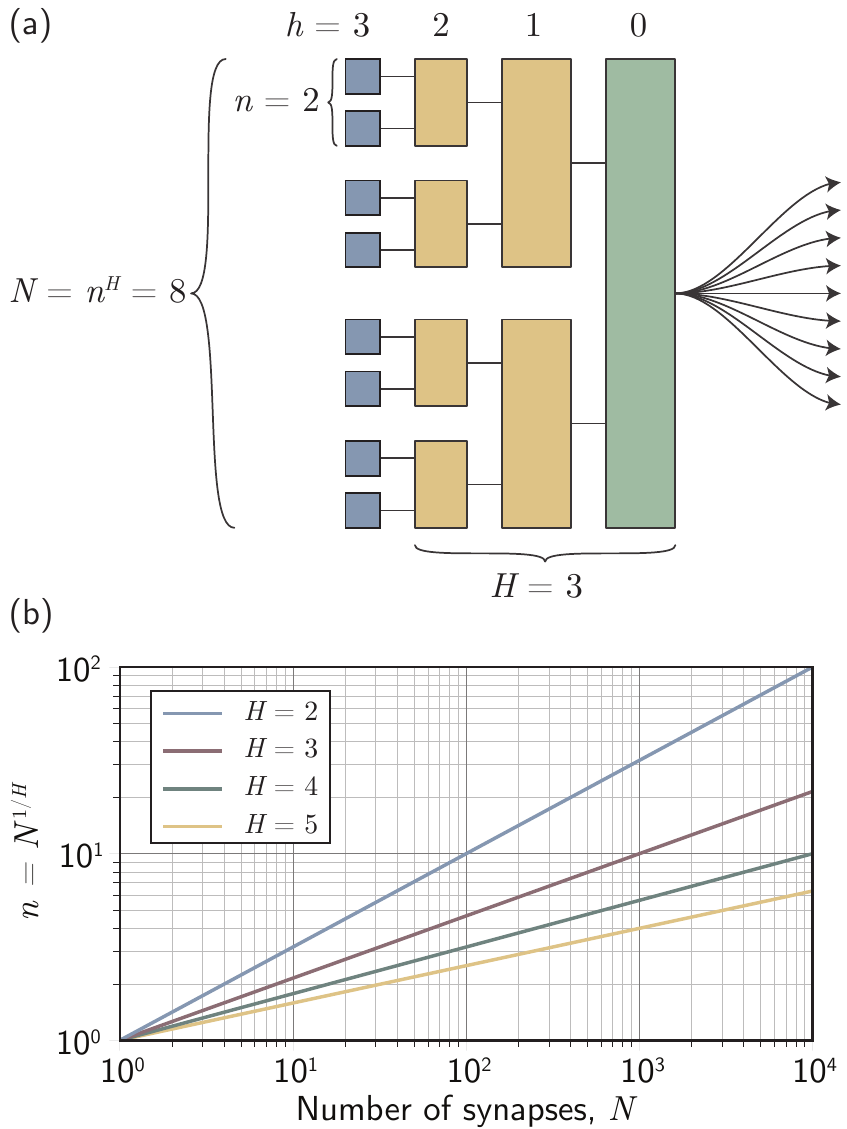}
\caption{Dendritic tree. (a) Schematic illustration of the tree structure with blue synapses input to yellow dendrites. The neuronal cell body is shown in green with fan-out to downstream synapses. The fan-in factor ($n$) is labeled, as is the hierarchy level ($h$), total depth of hierarchy ($H$), and the total number of synapses ($N$). (b) The fan-in factor as a function of the total number of synapses in a neuron for several values of the depth of the hierarchy. $N$, $n$, and $H$ must be integers if a homogeneous tree is used.}
\label{fig:dendritic_tree}
\end{figure}

A schematic of a dendritic tree is shown in Fig.\,\ref{fig:dendritic_tree}(a). The architecture consists of input synapses (shown in blue), multiple levels of dendritic hierarchy (yellow), and the final cell body (green). These components have been defined in Sec.\,\ref{sec:introduction}, and all three can be implemented with SQUID circuits, a self-similarity that facilitates scalable design and fabrication. A specific circuit implementation is given in Sec.\,\ref{sec:Loops}.

We restrict attention to a homogeneous dendritic tree of the form shown in Fig.\,\ref{fig:dendritic_tree}(a), wherein all dendrites receive the same number of inputs, $n$, which we refer to as the \textit{fan-in factor}. The neuron cell body resides at level zero of the dendritic hierarchy, and synapses reside at level $H$, so the total number of synapses is $n^H \equiv N$. In Fig.\,\ref{fig:dendritic_tree}(a) we show a tree with fan-in factor $n = 2$ and three levels of hierarchy for a total of $N = 8$ synapses. For a homogeneous dendritic tree, the relationship between number of synapses, fan-in factor, and hierarchy is plotted in Fig.\,\ref{fig:dendritic_tree}(b). Biological neurons are less uniform and more complex, but homogeneous trees should be a good starting point for artificial systems.

\begin{figure}[tb]
\centering
\includegraphics[width=8.6cm]{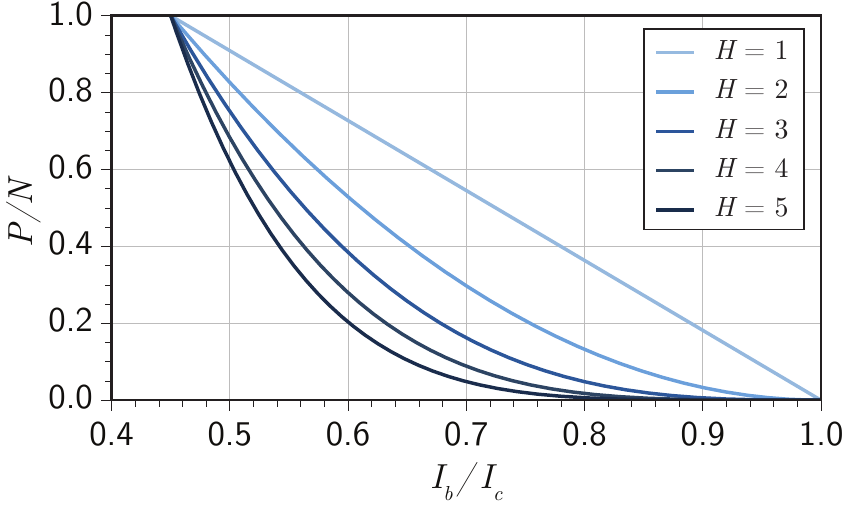}
\caption{The fraction of synapses required to be saturated to drive a neuron to threshold as a function of the normalized bias to dendrites and the cell body. This result depends on enforcing the condition that input flux is limited to $\Phi_0/2$ and that all $I_c$ values are identical, but the result is independent of $I_c$ and the total number of synapses in the neuron, $N$.}
\label{fig:results}
\end{figure}

Equation \ref{eq:fraction_active} is applicable to any dendrite or neuron cell body in the dendritic tree, provided the maximum applied flux is limited to $\Phi_0/2$. Working backward from the cell body, one can calculate that the minimum number of active synapses required to drive the neuron cell body to threshold is $P = p^H$, and the fraction of synaptic activity for threshold is at least
\begin{equation}
\label{eq:fraction_active__hierarchy}
\frac{P}{N} = \left( \frac{p}{n} \right)^H = \left[ \frac{3\pi + 2}{2\pi} \left( 1-\frac{I_b}{I_c} \right) \right]^H.
\end{equation}
Equation \ref{eq:fraction_active} is recovered as the special case of $H = 1$. The exponential dependence of threshold activity fraction on $H$ implies that dendritic trees can improve fan-in even with limited depth of the tree. This is illustrated in Fig.\,\ref{fig:results}, where the activity fraction as a function of bias is plotted for dendritic trees of varying depth. We see that the point neuron case ($H=1$) requires the highest activity fraction, but that the situation improves quickly with depth of dendritic tree. For instance, with $H=5$ and a conventional biasing of $0.7\,I_\mathrm{c}$, only 5\% of synapses need be active --- an order-of-magnitude improvement over the case of the point neuron. If $I_b/I_c$ can be pushed to $0.9$, a tree depth of only $H = 3$ is required to achieve sub-1\% threshold activity fraction. For biologically realistic fan-in of $10^4$ synapses, this dendritic tree would require dendrite fan-in of $n \approx 22$ and about 485 intermediate dendrites. Considering that every synapse requires a SQUID \cite{sh2020,khan2021modeling}, the additional hardware fraction for the dendritic tree is minor and area estimates of such hardware is found in appendix B of Ref.\,\cite{prsh2021}. These biological values are abjectly impossible for point neurons whose applied flux is limited to the range of monotonic response, providing a physical motivation for the use of dendritic trees in superconducting neurons.

In addition to the benefits of the sparse coding model, the lower activity fraction associated with an active dendritic tree also benefits the dynamic range of possible thresholds. Many homeostatic mechanisms involve the tuning of the bias point in response to network activity. Ideally, this bias point could be tuned over a wide range. By allowing lower bias points for the same activity fraction, dendrtic trees allow the threshold of a neuron to be tuned over a wider range than that of a point neuron.

The energy consumption of the dendritic arbor itself also deserves consideration. For future superconducting systems, dynamic power should dominate static power consumption. The total fraction of all units (synapses, dendrites, and soma) that must be active to reach threshold is given by
\begin{equation}
\label{eq:fraction_active__dendrites}
\frac{P_\mathrm{tot}}{N_\mathrm{tot}} = \frac{\sum_{h = 0}^H p^h}{\sum_{h = 0}^H n^h}.
\end{equation}

The energy consumption of synapses and dendrites is unlikely to be the same for most technologies. In the optoelectronic case, for example, synaptic events are likely to cost significantly more power than an active dendrite. Additionally, it can be shown that the total number of active units is likely to be higher in the point neuron case for almost all reasonable bias conditions as the number of added dendrites is compensated for by the greatly reduced number of active synapses.
%

\section{Loop Neuron Circuits}
\label{sec:Loops}
\begin{figure}[b!]
\includegraphics[width=8.6cm]{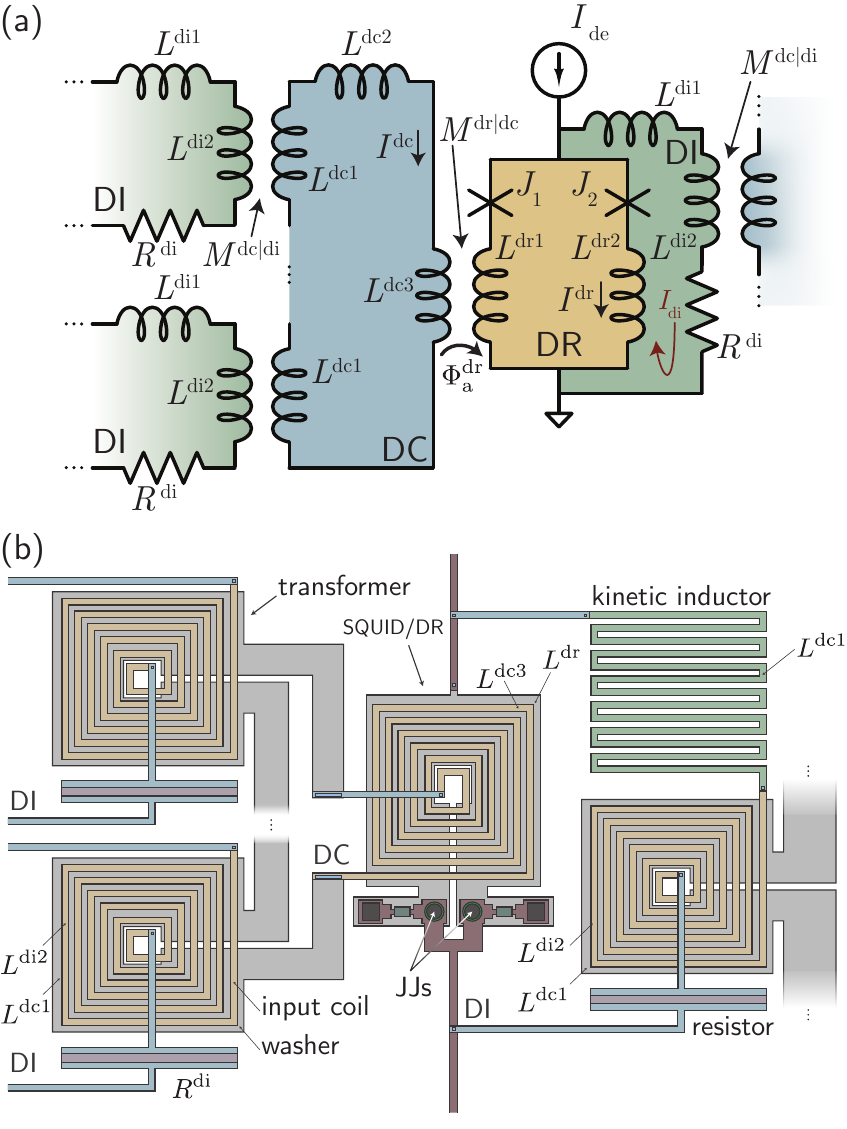}
\caption{(a) Circuit under consideration. Input dendritic integration (DI) loops couple signal into the dendritic collection (DC) loop via transformers. The net induced signal in the DC loop couples into the dendritic receiving (DR) loop, which is a SQUID. This SQUID is embedded in its own DI loop, which performs leaky integration on the accumulated signal. (b) Schematic of physical layout of circuit with components playing the roles of the circuit elements in (a). Circuit elements and loops are labeled to be consistent with the text.}
\label{fig:circuit}
\end{figure}
Neuronal circuits based on superconducting loops have been proposed in prior work, particularly with regard to optoelectronic systems \cite{sh2019}. We show here an application of these fan-in considerations to the specific case of loop neurons. A circuit diagram is shown in Fig.\,\ref{fig:circuit}. The dendritic integration (DI) loop integrates signals from activity present at that dendrite (or synapse). The saturation current of the DI loop corresponds to the maximally weighted active synapse discussed in the previous section. A mutual inductor ($M^\mathrm{dc|di}$) couples this signal into a second loop, called the dendritic collection (DC) loop. This loop is not strictly necessary, but allows for a more standardized design procedure, as discussed below. The DC loop applies flux $\Phi_\mathrm{a}^\mathrm{dr}$ (the weighted contribution of afferent signals) through $M^\mathrm{dr|dc}$ to the dendritic receiving (DR) loop. The DR loop forms the active component of the dendrite that has been the subject of our discussion thus far. Its output, a train of fluxons, is then coupled into another DI loop, allowing the chain to continue indefinitely. A schematic layout is provided in Fig.\,\ref{fig:circuit}\,(b) to provide physical intuition about the circuit. 

Still, the question remains: How do we limit the applied flux $\Phi_\mathrm{a}^\mathrm{dr}$ to enforce monotocity in practice? For this circuit, a careful choice of inductances is all that is required. The mathematical details are given in Appendices \ref{apx:squid} and \ref{apx:inductance_constraint}, but ultimately only a single constraint amongst all of the inductances is necessary (Eq.\,\ref{eq:inductance_constraint}). Additionally, the intermediate DC loop allows the monotonic condition to be met across a wide range of fan-in factors with only $L^{\mathrm{di2}}$ being a function $n$. This means that the SQUID and its input coil need not be redesigned for different choices of $n$. The consequences of the DC loop are further explored in Appendix\,\ref{apx:alternative_scenarios}.

\section{\label{sec:discussion}Discussion}
We have considered the implications of limiting the maximum flux input to all SQUIDs so the response is monotonically increasing. It is found that limiting the applied flux introduces a constraint on the activity fraction of synapses required to reach threshold, and the addition of a dendritic tree ameliorates the situation. This behavior is independent of most details of the circuit (such as whether or not a collection loop is used). The physical arguments presented here in favor of dendritic trees are derived from this decision to limit the applied flux to handle the ostensible ``worst-case'' scenario in which all synaptic inputs are fully saturated simultaneously. It is fair to question whether it is necessary to design our circuits around this extreme situation. The monotonicity issue could, for instance, be solved by immediately resetting all post-synaptic potentials to zero upon threshold. This is the standard behavior exhibited by most leaky integrate-and-fire models. However, implementing such a mechanism in superconducting hardware without compromising the speed and efficiency of superconducting neurons appears challenging. Additionally, we have argued elsewhere \cite{sh2020} that SQUID dendrites provide numerous opportunities for active, analog dendritic processing independent of the fan-in benefits described here. In that context, enforcement of monotonicity appears necessary. For these reasons, we contend that the best course of action is to allow synaptic signals to decay naturally without regard to thresholding events (which also preserves information) while limiting the applied flux in the manner described.

Still, one could argue we are over-preparing for the worst case scenario. Perhaps we could leave the maximum possible applied flux to each SQUID unrestricted, and instances wherein SQUIDs are driven past a half-period of their response function will be sufficiently rare that we can ignore them in design. For general cognitive activity, we are likely to seek networks balanced at a critical point \cite{stanley1999scaling,bata1987,bata1988} between excessive synchronization (order) and insufficient correlation (disorder). When cognitive circuits are poised close to this critical point, neuronal avalanches \cite{beggs2003neuronal} or cell assemblies \cite{plenz2007organizing,buzsaki2010neural} are observed to be characterized by a power-law \cite{be2007} or log-normal \cite{buzsaki2014log} distribution of sizes. A great deal of contemporary research \cite{tomen2019functional} indicates that operation near this critical point is advantageous for maximizing dynamic range \cite{kico2006,shya2009} and the number of accessible metastable states \cite{haldeman2005critical} while supporting long-range correlations in network activity \cite{kism2009}. With either power-law or log-normal distributions, network activity engaging many neurons is less probable than activity involving few neurons, but periods of activity involving large numbers of neurons are not so improbable as to be neglected and may be crucial episodes for information integration across the network. The probability of large events does not decay exponentially and must therefore be accommodated in hardware.

We reiterate that the primary assumption entering the derivation of Eq.\,\ref{eq:fraction_active} is that the maximum applied signal is limited to a certain value. We have considered the ramifications in the specific context of SQUID components, but similar considerations may apply to other hardware. We encourage the reader to consider whether similar arguments may affect their favorite neuromorphic thresholding elements. We also note that limiting the applied flux to $\Phi_0/2$ may not always be advisable. From the activation function of Fig.\,\ref{fig:squid_response}(b) it is evident a dendrite with two synapses performs XOR if each synapse couples $\Phi_0/2$ into the receiving SQUID. When both synapses are active, the device operates outside the monotonic response. We hope this article does not stifle the investigation of the full neural utility of engineered SQUID responses.

\section*{Acknowledgements}
We thank Dr. Ken Segall and Dr. Michael Schneider for helpful discussions.

\appendix
\section{\label{apx:squid}Typical SQUID design}
The inductance of the SQUID washer ($L^{\mathrm{dr}}$ in Fig.\,\ref{fig:circuit}(b)) is determined by the basic theory of operation \cite{clbr2006}, which dictates that
\begin{equation}
\label{eq:squid_beta_L}
\beta_{L} = \frac{ 2 L^{\mathrm{sq}} I_c }{\Phi_0} = 1 \ \longrightarrow \  L^{\mathrm{dr}} = L^{\mathrm{sq}} = \frac{\Phi_0}{ 2 I_c }.
\end{equation}
The total inductance of a SQUID is given by $L^{\mathrm{sq}}_{\mathrm{tot}} = L^{\mathrm{sq}} + L_{j1} + L_{j2}$. $L_{j1}$ and $L_{j2}$ are the inductances of the two JJs in the SQUID. The inductance of a JJ is a function of the instantaneous current through the junction and varies from $L_j(0) = \Phi_0/2\pi I_c$ at $I = 0$ to $(\pi/2)L_j(0)$ at $I = I_c$ (see Ref.\,\onlinecite{vatu1998}, pp. 210-211). When the SQUID is close to threshold, one junction will have $I \approx I_c$. The current through the other junction will depend on the value of the bias, but for simplicity we take the inductance value corresponding to $I = 0$. The total inductance of the SQUID (DR loop) is then given by
\begin{equation}
L^{\mathrm{sq}}_{\mathrm{tot}} = \frac{\Phi_0}{I_c} \left( \frac{3\pi+2}{4\pi} \right).
\end{equation}
This approximation will cause an error less than $\pi/2$ in the prefactor of Eq.\,\ref{eq:fraction_active}, and likely significantly less.

\section{\label{apx:inductance_constraint}Mathematical details in the case of a collection loop}
To ensure the maximum signal applied to a dendrite or neuron cell body is limited to $\Phi_0/2$, we must consider the specific circuit used to collect the signals. Here we consider the circuit of Fig.\,\ref{fig:circuit}(a) wherein a collection coil (DC loop) is used to sum input flux, and we derive a relation that must be satisfied by the inductors in the circuit.

The total applied flux to a DR loop due to arbitrary current in the input DI loops is referred to as $\Phi_{\mathrm{a}}^{\mathrm{dr}}$:
\begin{equation}
\label{eq:fan-in__transformer_collection__full}
\Phi_{\mathrm{a}}^{\mathrm{dr}} = \frac{ M^{\mathrm{dr|dc}} }{ L^{\mathrm{dc}}_{\mathrm{tot}} } \, \sum_{i=1}^{n} M^{\mathrm{dc|di}}_i I_i^{\mathrm{di}}.
\end{equation}
$I_i^{\mathrm{di}}$ is the current in the integration loop of the $i$th input to the dendrite at a given time, and the sum is over the $n$ inputs. In the case of a point neuron, the sum is over all $n$ synapses. The total inductance of the DC loop is given by 
\begin{equation}
\label{eq:inductance_dc_tot}
L^{\mathrm{dc}}_{\mathrm{tot}} = \sum_i L^{\mathrm{dc1}}_i + L^{\mathrm{dc2}} + L^{\mathrm{dc3}},
\end{equation} 
and the mutual inductors are 
\begin{equation}
\label{eq:mutual_inductances}
\begin{split}
M^{\mathrm{dc|di}} &= k_1 \sqrt{ L^{\mathrm{di2}} \, L^{\mathrm{dc1}} }, \\
M^{\mathrm{dr|dc}} &= k_2 \sqrt{ L^{\mathrm{dc3}} \, L^{\mathrm{dr}} },
\end{split}
\end{equation}
with $L^{\mathrm{dr}} = L^{\mathrm{dr1}}+ L^{\mathrm{dr2}}$. See Fig.\,\ref{fig:circuit}(a) for definitions of circuit parameters. 

Due to the stipulation that all dendrites and neurons will have a monotonic response function, circuit parameters are thus constrained so the maximum value of $\Phi_{\mathrm{a}}^{\mathrm{dr}}$ as given by Eq.\,\ref{eq:fan-in__transformer_collection__full} is less than or equal to $\Phi_0/2$. The behavior of the DI loop of the circuit in Fig.\,\ref{fig:circuit} is such that the integrated current cannot exceed a certain value, which we refer to as $I_{\mathrm{sat}}^{\mathrm{di}}$. This behavior is explained in Ref.\,\onlinecite{khan2021modeling}. For the present analysis, we restrict attention to homogeneous dendrites wherein all $M^{\mathrm{dc|di}}$ are identical, and all inputs have the same value of $I_{\mathrm{sat}}^{\mathrm{di}}$. We enforce monotonic response with the condition that
\begin{equation}
\label{eq:monotonic_response_condition}
\Phi_{\mathrm{max}}^{\mathrm{dr}} = n \, \frac{ M^{\mathrm{dr|dc}} \, M^{\mathrm{dc|di}}}{ L^{\mathrm{dc}}_{\mathrm{tot}} } \, I_{\mathrm{sat}}^{\mathrm{di}}.
\end{equation}

Inserting the expressions from Eqs.\,\ref{eq:inductance_dc_tot} and \ref{eq:mutual_inductances} into Eq.\,\ref{eq:monotonic_response_condition} and rearranging, we arrive at the constraint on the inductances in the circuit:
\begin{equation}
\begin{split}
\label{eq:inductance_constraint}
& L^{\mathrm{di2}} = \frac{1}{L^{\mathrm{dr}}} \times \\
& \left\{ \frac{\Phi^{\mathrm{dr}}_{\mathrm{max}}}{k_1 k_2 I^{\mathrm{di}}_{\mathrm{sat}}} \left[ \left( \frac{L^{\mathrm{dc1}}}{L^{\mathrm{dc3}}} \right)^{1/2} + \left( \frac{L^{\mathrm{dc3}}}{L^{\mathrm{dc1}}} \right)^{1/2} \frac{1+\alpha}{n} \right]  \right\}^2.
\end{split}
\end{equation}
We have simplified Eq.\,\ref{eq:inductance_constraint} without loss of generality by expressing the parasitic inductance $L^{\mathrm{dc2}}$ as a fraction of $L^{\mathrm{dc3}}$: $L^{\mathrm{dc2}} = \alpha L^{\mathrm{dc3}}$. The inductances must be chosen so that Eq.\,\ref{eq:inductance_constraint} is satisfied if the total flux is to be limited so the SQUID retains a monotonic response. Equation \ref{eq:inductance_constraint} is valid when a collection loop is used to accumulate signals, as shown in Fig.\,\ref{fig:circuit}. Analysis of the circuit without a collection loop is given in Appendix \ref{apx:alternative_scenarios}.

\begin{figure}[h!]
\centering
\includegraphics[width=8.6cm]{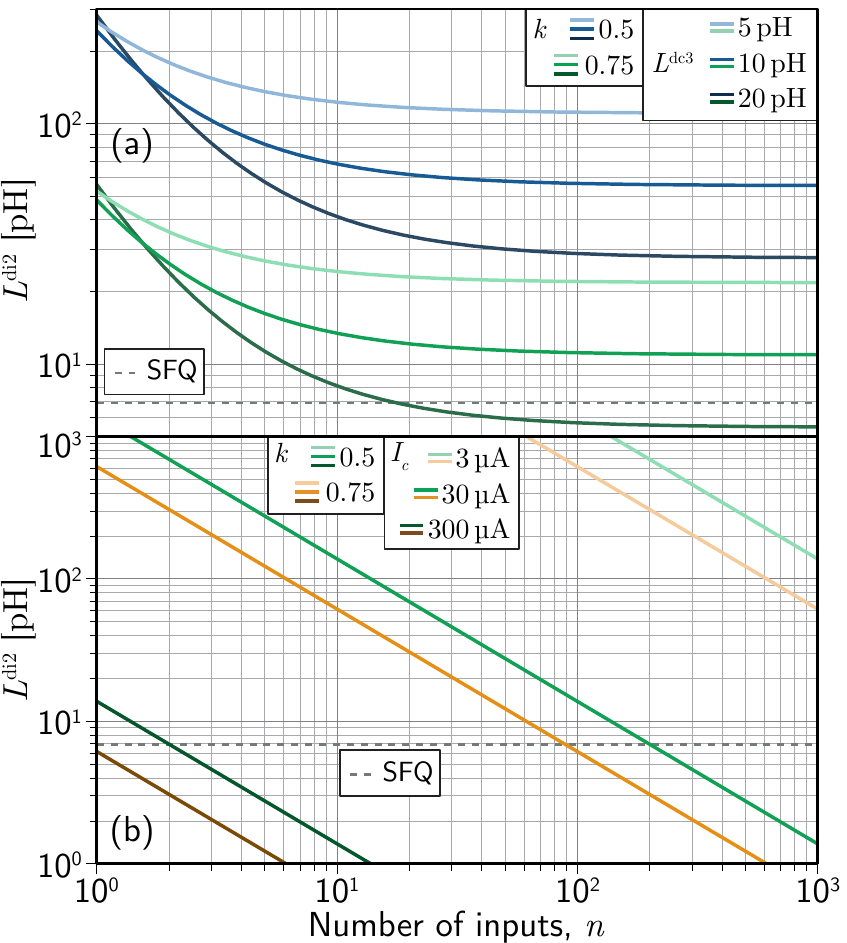}
\caption{(a) Inductance $L^{\mathrm{di2}}$ required to satisfy the constraint that the maximum applied flux is limited to $\Phi_0/2$. A collection loop is used satisfying the relationship of Eq.\,\ref{eq:inductance_constraint} with $I_c = 300$\,\textmu A, $\Phi^{\mathrm{dr}}_{\mathrm{max}} = \Phi_0/2$, $L^{\mathrm{dc1}} = 10$\,pH, and $k_1 = k_2$. The value $L^{\mathrm{di2}} = \Phi_0/I_c$ is labeled `SFQ', as this is the value of $L^{\mathrm{di2}}$ at which the DI loop saturates with a single flux quantum. (b) Inductance $L^{\mathrm{di2}}$ required to satisfy the constraint that the maximum applied flux is limited to $\Phi_0/2$ when no collection loop is used (Eq.\,\ref{eq:inductance_constraint__no_collection__further_specified}).}
\label{fig:Ldi2}
\end{figure}

Equation \ref{eq:inductance_constraint} can be applied to the case where synapses/dendrites store only single-flux quanta in their integration loops. This case corresponds to setting $L^{\mathrm{di2}} = \Phi_0/I_c$, and this line is shown in Fig.\,\ref{fig:Ldi2}. One finds that for many values of $n$ with this choice for $L^{\mathrm{di2}}$, Eq.\,\ref{eq:inductance_constraint} cannot be satisfied. The situation improves with increasing transformer efficiency $k$ and for larger $n$. 

With or without the collection loop, the inductance of the SQUID washer is determined based on the considerations given in Appendix \ref{apx:squid}. We envision a standard washer inductor with flux input of the form shown in Fig.\,\ref{fig:circuit}(b), which is why the total SQUID inductance enters the mutual inductance of the transformer given in Eq.\,\ref{eq:mutual_inductances}. $L^{\mathrm{dr}}$ is the inductance of the washer, while $L^{\mathrm{dc3}}$ is the inductance of the input coil. The inductance of the DR loop is set by Eq.\,\ref{eq:squid_beta_L}, but the washer inductors used to receive inputs in the DC loop ($L^{\mathrm{dc1}}$) do not face the same SQUID operation criterion. We choose $L^\mathrm{dc1} = 10$\,pH based on simulation results, but note that there is some flexibility in choosing this parameter, as only the ratio of $L^\mathrm{dc1}$ and $L^{\mathrm{dc3}}$ enters Eq.\,\ref{eq:inductance_constraint}. 

We next must choose a junction $I_c$. We have discussed elsewhere that the SQUID faces an area-energy trade-off \cite{prsh2021}. It was found that 300\,\textmu A achieves an acceptable compromise when seeking large-scale systems with photonic communication. We use this value to specify inductances, but we remind the reader that the value of $I_c$ does not affect the fraction of activity required for threshold given by Eq.\,\ref{eq:fraction_active__hierarchy}. $I_c$ does determine $I^{\mathrm{di}}_{\mathrm{sat}}$, and we make use here of a relation of the form $I^{\mathrm{di}}_{\mathrm{sat}} = \gamma I_c$, with $\gamma = 1$ used throughout.

The inductances of the input coils ($L^{\mathrm{di2}}$ and $L^{\mathrm{dc3}}$) then remain to be specified. The self-inductances of the DI loops do not enter the present consideration, although they do affect cross-talk, discussed below. It is our present understanding that the values of $L^{\mathrm{di2}}$ and $L^{\mathrm{dc3}}$ can be chosen with some flexibility, provided Eq.\,\ref{eq:inductance_constraint} is satisfied. In hardware, these parameters are determined by the geometry of the input coil in a washer configuration like that of Fig.\,\ref{fig:circuit}(b) (wire width and pitch, number of turns). Typical values are 10\,pH - 1\,nH. In the present context, we are primarily interested in analog DI loops, wherein the total inductance of the DI loop is $L^{\mathrm{di}}_{\mathrm{tot}} = L^{\mathrm{di1}} + L^{\mathrm{di2}}$ with $L^{\mathrm{di1}} \gg L^{\mathrm{di2}}$, so the value of $L^{\mathrm{di2}}$ can typically be chosen without affecting the analog circuit parameters of the DI loop.

Equation \ref{eq:inductance_constraint} is plotted in Fig.\,\ref{fig:Ldi2}\,(a) as a function of the number of inputs to the dendrite or neuron, $n$, for two values of mutual inductor efficiency $k_1 = k_2 = k$. For a smaller number of inputs, the input coil of each DI loop must be larger to overcome the inertia of $L^{\mathrm{dc3}}$ in the collection coil, an effect that increases with increasing $L^{\mathrm{dc3}}$. The required value of $L^{\mathrm{di2}}$ approaches an asymptotic value with $n$, and in the large-$n$ limit $L^{\mathrm{di2}}$ scales inversely with $L^{\mathrm{dc3}}$. 
InductEx simulations \cite{fourie2011three}, confirmed that all of the points on the $k = 0.5$ (blue) curves in Fig.\,\ref{fig:Ldi2}\,(a) can be fabricated to fit within an area of 20\,\textmu m $\times$ 20\,\textmu m. Simulations assumed a traditional square washer and input coil fabricated from Niobium with a spacing of 100\,nm and a minimum feature size of 100\,nm. Other geometries could be explored to further reduce the size of these circuit components (a necessary development to enable the scaling described in Ref.\,\onlinecite{prsh2021}.) While the equations given here can guide the design of future circuits, determining the optimal parameters for all circuit elements will be a complicated decision based on energy/size trade-offs of SQUIDs \cite{prsh2021}, fabrication technology, cross-talk, and the ability to interface with other sub-circuits (such as optical transmitters or receivers for the optoelectronic case). As such, the parameters given in this appendix should not be interpreted as the optimal operating point for all future systems.

In the main text, we derived Eq.\,\ref{eq:fraction_active__hierarchy} without reference to many specific circuit details. However, we can derive the same general expression in the specific circuit context under consideration. A dendrite's threshold current ($I^{\mathrm{dr}}_{\mathrm{th}}$) is reached when the sum of the bias current ($I_b$ provided by $I^{\mathrm{de}}$ split between the two JJs of the SQUID) and the current induced by input flux ($I_{\mathrm{a}}^{\mathrm{dr}}$) equal the junction critical current, $I_c$:
\begin{equation}
\label{eq:current_threshold}
I^{\mathrm{dr}}_{\mathrm{th}} = I_c - I_b.
\end{equation} 
The applied current due to input flux is given by
\begin{equation}
\label{eq:I_dr_applied}
I^{\mathrm{dr}}_{\mathrm{a}} = \frac{\Phi^{\mathrm{dr}}_{\mathrm{a}}}{L^{\mathrm{dr}}_{\mathrm{tot}}},
\end{equation}
where $L^{\mathrm{dr}}_{\mathrm{tot}}$ has been introduced in Appendix \ref{apx:squid}. 

As was used above to constrain the input inductances, the maximum applied signal to the dendrite results when all $n$ inputs have current $I^{\mathrm{di}}_{\mathrm{sat}}$. We now ask what number $p$ of inputs must be driven to their saturated value to drive a dendrite to threshold. Using Eq.\,\ref{eq:fan-in__transformer_collection__full} with Eq.\,\ref{eq:I_dr_applied}, Eq.\,\ref{eq:current_threshold}, and inserting derived expressions for all inductances, we can calculate the quantity $p/n$, which gives the 
fraction of total possible input activity that is sufficient to make the dendrite active. We find the following result.
\begin{equation}
\label{eq:fraction_active__apx}
\frac{p}{n} = \frac{3\pi + 2}{2\pi} \left( 1-\frac{I_b}{I_c} \right).
\end{equation}
This expression is identical to Eq.\,\ref{eq:fraction_active} derived in the main text with minimal consideration of the specifics of the circuit. The important point is that the activity fraction driving a dendrite above threshold is independent of $n$ and depends only on how closely the dendrite is biased to its critical current. Equation \ref{eq:fraction_active__apx} holds independently of whether or not a collection coil is used, and results only from the constraint to limit the applied flux to $\Phi_0/2$. The approximation of the JJ inductances determined the prefactor in Eq.\,\ref{eq:fraction_active__apx}, but does not affect the functional form. Equation \ref{eq:fraction_active__apx} also holds whether the SQUID junction critical currents are the same or different than those used in the synapses, and is independent of the value of $L^{\mathrm{di1}}$, so this condition applies to single-flux or analog synapses.

We would also like to know how much current is induced in a given DI loop when appreciable signal is accumulated in the other loops connected to the same collection coil. Using the same design choices and inductances given above, we arrive at the following expression for induced current in a DI loop when a fraction $p$ of inputs is at their saturation capacity:
\begin{equation}
\label{eq:cross_talk}
I^{\mathrm{di}}_{\mathrm{ind}} =  p \ \frac{k_1^2 L^{\mathrm{di2}} L^{\mathrm{dc1}} }{L^{\mathrm{dc}}_{\mathrm{tot}} L^{\mathrm{di}}_{\mathrm{tot}}} \ I^{\mathrm{di}}_{\mathrm{sat}}.
\end{equation}

\section{\label{apx:alternative_scenarios}Scenarios without a collection loop, allowing $I_c$ to vary, and seeking single-flux operation}
If no DC loop is used to collect inputs, the applied flux to the SQUID comprising the DR loop is given by
\begin{equation}
    \label{eq:applied_flux__no_collection}
    \Phi^{\mathrm{dr}}_{\mathrm{a}} = \sum_{i = 1}^{n}M_i^{\mathrm{dr|di}}I_i^{\mathrm{di}}.
\end{equation}
In this case, we assume the SQUID washer is separated into $n$ discrete inductive elements comprising the $n$ transformers, satisfying the condition
\begin{equation}
    \label{eq:inductance_dr_tot}
    \sum_{i = 1}^{n} L_i^{\mathrm{dr1}} \frac{\Phi_0}{2I_c} = n L^{\mathrm{dr1}}, 
\end{equation}
which results from the $\beta_L$ criterion of Appendix \ref{apx:squid}. For the case wherein a collection loop is not used is, the expression analogous to Eq.\,\ref{eq:inductance_constraint} is
\begin{equation}
\label{eq:inductance_constraint__no_collection}
L^{\mathrm{di2}} = \frac{1}{L^{\mathrm{dr1}}} \left( \frac{\Phi^{\mathrm{dr}}_{\mathrm{max}}}{n k I^{\mathrm{di}}_{\mathrm{sat}}} \right)^2.
\end{equation}
Equation \ref{eq:inductance_constraint__no_collection} must be satisfied if the total applied flux is to be limited to $\Phi^{\mathrm{dr}}_{\mathrm{max}}$. If we further specify $\Phi^{\mathrm{dr}}_{\mathrm{max}} = \Phi_0/2$ and take $I^{\mathrm{di}}_{\mathrm{sat}} = I_c$, as was assumed in the case of the collection loop in Sec.\,\ref{sec:model}, we arrive at
\begin{equation}
    \label{eq:inductance_constraint__no_collection__further_specified}
    L^{\mathrm{di2}} = \frac{\Phi_0}{2 n k^2 I_c}.
\end{equation}
One interpretation of Eq.\,\ref{eq:inductance_constraint__no_collection__further_specified} is that if no collection loop is used and the total applied flux is limited to $\Phi_0/2$, then the inductance of the DI loops ($L^{\mathrm{di2}}$) or the mutual inductance coupling factor ($k$) must become very small as $n$ grows. 

Equation \ref{eq:inductance_constraint__no_collection__further_specified} is plotted in Fig.\,\ref{fig:Ldi2}(b) for several values of junction $I_c$ and coupling $k$. It is assumed the junctions in the DI loops input to the SQUID and the junctions within the DR loop forming the SQUID have the same $I_c$. This plot shows that if the collection loop is omitted and typical values of $k$ are used, it may be necessary for the value of $I_c$ to vary from neuron to neuron depending on the number synapses input to that neuron. In this model, it is still assumed that the synapses and the neuron have the same $I_c$. When the collection loop is omitted, not only does the design of SQUID inductors become fan-in dependent, $L^\mathrm{di2}$ is also inversely proportional to $n$ for all $n$. Fabricating sub-picohenry inductors is difficult and may become a limiting factor without a collection loop.

If one also seeks single-flux operation wherein the signal from each synapse to a dendrite or neuron consists of a single magnetic flux quantum, it is necessary to limit the inductance to
\begin{equation}
    \label{eq:inductance_constraint__no_collection__sfq}
    L^{\mathrm{di2}} = \frac{\Phi_0}{I_c}.
\end{equation}
Equations Eq.\,\ref{eq:inductance_constraint__no_collection} and \ref{eq:inductance_constraint__no_collection__sfq} can both be satisfied if $k = (2n)^{-1/2}$. Thus, if no collection coil is used and single-flux operation is employed without changing the junction $I_c$ between the DI loops and the DR loop, the coupling $k$ must decrease as one over the square root of $n$, becoming quite small for large $n$.

Another option is to allow the junction $I_c$ to differ between the junctions in the DI loop and the DR loop. In this case, requiring the maximum applied flux be limited to $\Phi_0/2$ results in the condition
\begin{equation}
\label{eq:inductance_constraint__no_collection__vary_Ic}
L^{\mathrm{di2}} = \frac{\Phi_0}{2 n k^2} \frac{I_c^{\mathrm{dr}}}{\left( I_c^{\mathrm{di}} \right)^2}.
\end{equation}
This condition has multiple free parameters, so it can be satisfied for arbitrary $n$. However, in diverse networks wherein different neurons have different numbers of inputs, the requirement of designing and current-biasing junctions with appreciably different critical currents will be cumbersome. If one also requires single-flux operation in the integration loops, the condition
\begin{equation}
    \label{eq:Ic_constraint__no_collection__sfq}
    I_c^{\mathrm{di}} = \frac{1}{nk^2} I_c^{\mathrm{dr}}
\end{equation}
must be satisfied. For large $n$, the junction $I_c$ of the inputs must become very small relative to the receiving SQUID. This condition results from limiting total applied flux to $\Phi_0/2$ while also requiring integration loops saturate with a single flux quantum. By contrast, Ref.\,\onlinecite{schneider2020fan} moved in the other direction and reduced the $I_c$ of the receiving loops relative to the input synapses to reduce the number of active synapses capable of driving the neuron to threshold. However, that work did not enforce that the maximum possible applied flux be limited to ensure a monotonic response, in part because a primary application of interest was for single-flux signals wherein no temporal integration is performed.

One can combine the expressions of this Appendix to find that when no collection loop is used, the $I_c$ of the DI loop junction and the DR loop junctions are allowed to take arbitrary values, and whether or not single-flux operation is pursued, the fraction of saturated synapses necessary to drive a point neuron to threshold is given by Eq.\,\ref{eq:fraction_active}. For high fan-in neurons, it is not possible to limit the applied flux to $\Phi_0/2$ and simultaneously enable threshold to be reached with a small number of synapses without a dendritic tree. The active fraction is a general result of limiting $n$ identical total inputs to a maximum value, as described in Sec.\,\ref{sec:discussion}. It does not depend on the choices to use a collection coil or to fix the junction $I_c$ of all components.

\bibliographystyle{unsrt}
\bibliography{main}
\end{document}